\documentclass{article}

\usepackage{microtype}
\usepackage{graphicx}
\usepackage{subfigure}
\usepackage{booktabs} 
\usepackage{amsmath}
\usepackage{amsfonts}
\usepackage{amssymb}
\usepackage{hyperref}
\usepackage{dsfont}

\usepackage[accepted]{mlsys2021}

\mlsystitlerunning{TinyRL for Quadruped Locomotion}

\begin{document}

\twocolumn[
\mlsystitle{Tiny Reinforcement Learning for Quadruped Locomotion using Decision Transformers}



\mlsyssetsymbol{equal}{*}

\begin{mlsysauthorlist}
\mlsysauthor{Orhan Eren Akg\"un}{equal,to}
\mlsysauthor{Néstor Cuevas}{equal,to}
\mlsysauthor{Matheus Farias}{equal,to}
\mlsysauthor{Daniel Garces}{equal,to}
\end{mlsysauthorlist}

\mlsysaffiliation{to}{Harvard University}

\mlsyscorrespondingauthor{Daniel Garces}{dgarces@g.harvard.edu}
\mlsyscorrespondingauthor{Orhan Eren Akgün}{erenakgun@g.harvard.edu}

\mlsyskeywords{Machine Learning, MLSys}

\vskip 0.3in

\begin{abstract}
Resource-constrained robotic platforms are particularly useful for tasks that require low-cost hardware alternatives due to the risk of losing the robot, like in search-and-rescue applications, or the need for a large number of devices, like in swarm robotics. For this reason, it is crucial to find mechanisms for adapting reinforcement learning techniques to the constraints imposed by lower computational power and smaller memory capacities of these ultra low-cost robotic platforms. We try to address this need by proposing a method for making imitation learning deployable onto resource-constrained robotic platforms. Here we cast the imitation learning problem as a conditional sequence modeling task and we train a decision transformer using expert demonstrations augmented with a custom reward. Then, we compress the resulting generative model using software optimization schemes, including quantization and pruning. We test our method in simulation using Isaac Gym, a realistic physics simulation environment designed for reinforcement learning. We empirically demonstrate that our method achieves natural looking gaits for Bittle, a resource-constrained quadruped robot. We also run multiple simulations to show the effects of pruning and quantization on the performance of the model. Our results show that quantization (down to 4 bits) and pruning reduce model size by around 30\% while maintaining a competitive reward, making the model deployable in a resource-constrained system.
\end{abstract}
]



\printAffiliationsAndNotice{\mlsysEqualContribution} 

\section{Introduction}
\label{introduction}
In this paper, we focus on bridging the reality gap of imitation learning for quadruped locomotion on resource-constrained robots. The reality gap refers to the inability of some learning algorithms to be deployed in real systems due to their computational or memory requirements, or the reliance on sensors and information that are not present in real-life systems \cite{Salvato2021}. This problem is particularly relevant when considering resource-constrained devices, as such devices tend to have cheaper, less accurate hardware, which leads to a mismatch between the memory, computational power, and information required to run the algorithms and the resources actually available in the real system. We are particularly interested in the reality gap for imitation learning algorithms, a subclass of reinforcement learning algorithms that leverage expert demonstrations to teach a learning agent a policy that approximates the desired behavior. 

During the last decade, imitation learning algorithms have seen major successes on complex tasks, including quadruped locomotion using animal motion as a reference \cite{Peng2020LearningAR}, driving tasks using expert input for refining training data \cite{ross2011reduction}, and other locomotion and manipulation tasks derived from observations alone (without being provided with a reward signal from the expert) \cite{kidambi2021mobile} \cite{sun2019provably} \cite{ho2016generative}. All of these works targeted either completely simulated environments or high-end robotic platforms, relying on the robots' vast computational resources, memory capabilities, and precise sensors and actuators to successfully execute the learned policies. This dependence on expensive hardware that is not easily accessible prevents the deployment of such algorithms to ultra-low-cost, resource-constrained robots. Adapting such algorithms so that they can be deployed to resource-constrained robots is important since these smaller devices could be used for tasks that require lower-cost hardware alternatives due to the risk of losing the robot, like in search and rescue applications, or the need for a large number of smaller devices, like in swarm robotics \cite{neuman2022tiny}. Recent research efforts on constrained robotics have achieved fall recovery and jumping skills in simulation using model-free reinforcement learning \cite{Zhang2021}, quadruped locomotion using linear policies for both simulation and hardware \cite{Rahme2021}, and motion using standard imitation learning techniques \cite{jabbour2022closing}. However, to the extent of our knowledge, there has been no success on bridging the reality gap for imitation learning of natural-looking quadruped gaits on resource-constrained devices.

For this reason, we try to bridge this gap by proposing a method that leverages expert demonstrations to produce natural-looking gaits, even with limited memory, sensor access, and computing power. Our method is based on decision transformers \cite{Chen2021DecisionTR}, a framework that abstracts reinforcement learning as a sequence modeling task. We use this framework to cast the imitation learning problem as a sequence-to-sequence learning task, using expert demonstrations and our own arbitrarily designed reward to conditionally learn a sequence of state-action pairs that will result in the desired motion. Since decision transformers utilize a version of GPT-2 \cite{Radford2019LanguageMA}, the resulting sequence modeling architecture has over 100M parameters, making it undeployable in a memory-constrained device. To solve this issue, we apply a series of software optimizations, including quantization and pruning \cite{han2016deep_compression}, to effectively decrease the model's size and allow its deployment in a low-cost, memory-constrained robotic platform. 

Hence, our contributions are as follows: 
\begin{itemize}
    \item We adapt the Decision Transformers to the task of quadruped locomotion in the Bittle platform;
    \item We formulate a custom reward structure to augment expert demonstrations and allow their use in the training of the adapted decision transformer;
    \item We leverage quantization and pruning strategies to compress the resulting sequence modeling architecture and allow its deployment in a real resource-constrained robotic platform;
    \item We empirically demonstrate that our method results in natural-looking gaits in simulation and enables real-system deployment as future work.
\end{itemize}

The rest of the paper is organized as follows: Section \ref{sec:related_works} provides a brief overview of related works in imitation learning, sequence modeling, sim-to-real transfer, and software optimizations. In Section \ref{sec:problem_formulation}, we present the problem formulation for quadruped locomotion as a sequence modeling task and the problem formulation for reducing the size of the resulting sequence modeling architecture through software optimizations. In Section \ref{sec:challenges_results}, we present our experimental setup, results, and the challenges faced while developing our method. In Section \ref{sec:dicussion_future_work}, we discuss the overall implications of our results and future directions that we are considering for our work. \footnote{The code for this paper can be found at: \href{https://github.com/DAGABO98/DT-Quadruped-Locomotion}{https://github.com/DAGABO98/DT-Quadruped-Locomotion}}

\section{Related Works}
\label{sec:related_works}
The development of controllers for quadruped locomotion has been a focus of multiple research efforts during the last two decades. While some authors propose utilizing a variety of control strategies to generate locomotive skills \cite{Geyer2003} \cite{Bledt2018MITC3}, others have focused on endowing systems with learning and optimization capabilities to develop general locomotion policies \cite{Coros2011}. Following this distinction, we want to focus on learning algorithms, rather than on optimization and control approaches. In the following four subsections, we will provide more details for related works in imitation learning, sequence modeling for RL, sim-to-real transfer, and software optimizations for deployment of deep learning techniques.

\subsection{Imitation Learning}
Complex behaviors that would be hard to manually encode into controllers can be easily learned by using expert demonstrations paired with proper reward signals. Some early works in the field propose learning from animal gaits using motion tracking data \cite{Peng2020LearningAR}. Other works have focused on directly cloning specified behaviors \cite{torabi2018behavioral} and leveraging experts in the loop to generate more robust policies \cite{ross2011reduction}. While these works have been successful at generating natural looking gaits, they rely on experts directly providing reward signals. This strategy is not compatible with resource constrained robots that do not have encoders in their actuators and hence cannot generate a reward for tracking a motion. Other works try to solve this issue by learning from observations alone and casting the problem as a series of two-player min-max games \cite{sun2019provably} \cite{kidambi2021mobile}. Our proposed method lays at the intersection between imitation learning using expert given reward signals and the latter work of learning from observations alone. Here, we only consider expert demonstrations and later augment them using our own reward structure.

\subsection{Sequence Modeling for Reinforcement Learning (RL)}
Due to the conditional relationship between sequences of states and sequence of actions, sequence modeling techniques have found their way into RL applications. Most sequence modeling algorithms were originally developed for language tasks, including next word prediction \cite{Soam2022}, and text classification \cite{abdullahi2021deep}. Advances in these areas, however, have enabled the expansion of sequence modeling to other fields, including reinforcement learning. In \cite{Chen2021DecisionTR}, the authors introduce decision transformers, casting the problem of RL as a sequence modeling task conditioned on the reward to go. \cite{janner2021offline}, also casts the problem as a sequence modeling task by introducing trajectory transformers, but instead of predicting future actions, a trajectory transformer predicts sequences of future states. 
Since we are interested in obtaining a policy for controlling a quadruped robot, decision transformers provide a better backbone for our application.

\subsection{Sim-to-real transfer}
Sim-to-real transfer refers to the process of training a policy in simulation and then deploying it in a real-life system. This process can be facilitated by constructing better simulated environments \cite{Xie2020} \cite{tan2018sim}, and including real-world data sources for evaluation \cite{desai2020stochastic} \cite{hwangbo2019learning}. However, these methods only address some of the challenges, as real-life systems are usually plagued with noisy sensors and low-accuracy actuators, especially when dealing with ultra low-cost, resource-constrained platforms. To address this issue, simulators have incorporated domain randomization and noise schemes for the simulated environments \cite{chebotar2019closing} \cite{sadeghi2016cad2rl} \cite{tobin2017domain} \cite{peng2018sim}, leading to more robust and better suited policies for deployment in real systems. To tackle resource constraints while still making models as accurate as possible, other software optimizations have been introduced. In this work, we are interested in making our system deployable by leveraging such compression schemes, as explained in the next subsection.

\subsection{Software Optimizations}


Developing machine learning algorithms that run on resource-constrained, low-cost devices is an emerging field of machine learning \cite{shafique2021tinyml}. Resource-constrained robotics applications come with additional challenges because of the sensor and actuator limitations \cite{neuman2022tiny}. From a general tiny machine learning perspective, previous literature has explored how to decrease the model size of convolutional neural networks without significantly harming the accuracy \cite{han2015learning, han2016deep_compression}. These canonical approaches served as inspiration for further applications in multi-modal learning problems with transformers such as only-quantization strategies \cite{liu2021posttraining, ding2022, lin2022fqvit, bondarenko-etal-2021-understanding}, only pruning \cite{Zhu2021VisualTP, jiaoda2021}, other techniques different from quantization and pruning \cite{pmlr-v139-touvron21a, zong2021self, hao2022learning, yang2022nvit}, and also unified methods \cite{yu2022unified}.

Other authors also report state-of-the-art model compression techniques specifically applied to reinforcement learning problems \cite{su2021, dor2020, Lee2021GSTGT}. Following these recent successes, we further investigate quantization techniques and pruning, specifically applied to Decision Transformers for quadruped locomotion.

\section{Problem Formulation}
\label{sec:problem_formulation}
In this paper we are interested in learning a competitive sub-optimal policy based on expert demonstrations, using a custom reward signal assigned to the demonstration post-collection. In the following subsections we explain the formulation of expert demonstrations, their augmentation using a custom reward signal, the learning set-up, and the compression schemes used to make the proposed method deployable into a resource-constrained device.

\subsection{Expert demonstrations and custom reward structure}
\label{subsec:expert_demonstrations}
We assume expert demonstrations follow a Markov decision process (MDP) described by the tuple $(\mathcal{S}, \mathcal{A}, P)$. The MDP tuple is composed of states $s \in \mathcal{S}$, actions $a \in \mathcal{A}$, and transition dynamics $P(s'|s,a)$. Since experts are completely decoupled from the robotic platform in which the policy will be deployed, they do not specify a reward for a specific state-action pair. Instead, we augment each state-action pair with a reward specifically tailored for the targeted platform. We denote that reward as $r$ and the function used to generate such reward as $\mathcal{R}$. The function $\mathcal{R}$ only considers available sensors in the target platform and assign reward coefficients for each sensor based on motion targets known to be useful for quadruped locomotion \cite{Zhang2021}. The function used for our experiments is defined in Section \ref{sec:challenges_results}. For each time step $t$, we specify a triplet composed of a given action $a_t$, the current state $s_t$ at time $t$, and the incurred reward $r_t=\mathcal{R}(s_t,a_t)$ after executing action $a_t$ from state $s_t$. A single expert demonstration is expressed as a trajectory $\tau$, a sequence of states, actions, and rewards. We formally define a trajectory as $\tau = (s_0, a_0, r_0, s_1, a_1, r_1, \dots, s_H, a_H, r_H)$, where $H$ corresponds to the end of the time horizon. The total reward obtained for a given trajectory is then defined as $R = \sum_{t=0}^H r_t$. We create a set of training demonstrations $\mathcal{D}$ by collecting multiple expert demonstrations with varying degrees of success on the task of quadruped locomotion. The idea of augmenting expert observations with a custom reward structure allows us to have more control over which policies we want our system to learn and how relevant specific behaviors are to the successful execution of those policies. For this reason, we design the reward structure in such a way that the desired policies that we want to learn always achieve the maximum reward possible out of all the policies contained in the training data.

\subsection{Sequence modeling for quadruped locomotion}
For this problem, we are interested in the task of learning a competitive sub-optimal policy based on expert demonstrations. 
Using the standard reinforcement learning formulation, the objective of our method is to learn a policy that maximizes the expected return $\mathds{E}\left[ \sum_{t=1}^H r_t \right]$ by utilizing the expert demonstrations contained in the training data as defined in subsection \ref{subsec:expert_demonstrations}. We base our method on the decision transformers proposed in \cite{Chen2021DecisionTR}, which leverage the conditional modeling capabilities of GPT-2 to generate a sequence of future actions based on desired future returns (reward to go) and a sequence of $K$ previous states and actions. $K$, in this scenario, refers to the input window length and it is a hyperparameter of the algorithm that will be specified in Section \ref{sec:challenges_results}. The architecture proposed in \cite{Chen2021DecisionTR} is particularly advantageous for our scenario, as it allows to condition the generative process of GPT-2 on expected future rewards, allowing better performing policies to be preferred over faulty and inefficient policies. We deviate from the original formulation specified for decision transformers, as we do not allow for negative rewards and we allow individual rewards for action-state pairs to be floating point numbers of 32 bits in precision (for the uncompressed version of the model).

\subsection{Compressing the model}

Two of the most performant compression techniques reported in literature are pruning and quantization \cite{han2016deep_compression}. These two strategies have previously shown reasonable size reduction for neural networks at low accuracy cost, which aligns with our main goal in the context of robotic platforms. Therefore, we decided to study the implementation of quantization and pruning techniques on decision transformers. For example, we focused our pruning method only on the fully connected and convolutional layers found in the decision transformer architecture to approximate the approach to the canonical method described in \cite{han2016deep_compression}. Then, we further evaluate the size reduction considering that these layers are only a portion of the entire model.

Our model compression study is divided into two techniques: quantization and pruning.

For the quantization step, we formulate two questions:
\begin{enumerate}
    \item Which quantization function will be used to quantize?
    \item Which format are we going to target to represent our model parameters?
\end{enumerate}

For the first question, we must discuss the nature of the distribution of weights and activations in a neural network model. After training a model, we observe that the weights and activations follow a bell-shaped distribution \cite{lin2016, han2016deep_compression, pwlq}. As we want to determine clustering groups to round to the nearest value, ideally, we should map parameters following their distributions. However, to the best of our knowledge, there is no mathematical formulation of these distributions, thus, we cannot determine the optimal quantization function. Instead, the standard approach to deal with this question is to consider a uniform quantization scheme, where we divide the range of possible parameter values into intervals of the same length. Many prior works have studied uniform quantization, and despite it is not the ideal solution, it shows to be robust even for transformer models as referenced in the related works section.

The second question is likely to entail more engineering than cognitive work. There is no correct answer for the right number of bits to target after quantization. However, we always want to use the lowest number of bits possible to still achieve tiny and performant models. If our machine-learning pipeline is easily reproducible, previous literature often runs experiments with different numeric formats and analyzes the memory size reduction against loss in accuracy (see related works). Therefore, we evaluate the trade-off and then select the one with the best performance. Quantization seems to be a better compressing method compared to others, as aggressive size reductions of 4$\times$, or even 8$\times$, are not as detrimental for accuracy.

Pruning, however, is much more harmful to model accuracy, and we also have two preliminary questions to answer:
\begin{enumerate}
    \item How to determine the pruning policy?
    \item How to balance unstructured and structured pruning?
\end{enumerate}

The pruning policy question follows the same idea of the quantization function. Determining if a given neuron or connection is relevant to the model is a non-trivial problem, as it also depends on the model architecture. There is no exact metric to predict the impact of removing a connection without running inference with and without pruning it, which is an infeasible combinatorial problem (considering all the model connections). To address this question, we split the pruning technique into two steps: unstructured and structured pruning. In unstructured pruning, we consider connections individually as they are single elements for each weight matrix. Thus, we calculate the $L1$-norm of the connection, i.e., its magnitude. If the magnitude is below a given threshold, we prune it (we can also determine the percentage of connections we want to prune by sorting the least $L1$-normed connections). On the other hand, structured pruning considers neurons as the smallest element, so pruning structurally means removing all the connections between a neuron and its previous layer, i.e., removing the entire neuron from the model. There are many metrics to evaluate which neurons to drop, but here we will use the $L2$-norm of the vector corresponding to the connections attached to the neuron. To further understand different metrics to perform pruning see these comprehensive surveys \cite{LIANG2021370, xu2020, deng2020}.

The second question is also more engineering work. In fully connected layers, we represent connections with just a vector, so performing unstructured or structured pruning has the same effect. However, in convolutional layers, the connections are represented with matrices, so while unstructured pruning removes single elements, structured pruning is more harmful because it removes entire rows. The trade-off evaluation to balance these methods is that, besides being harsher, structured pruning has a direct impact on the model size reduction (removing neurons, filters, and channels decreases matrices dimensions), whereas unstructured just increases sparsity. Clearly, if the sparsity is very pronounced, we can use different algorithms to perform sparse matrix multiplication and also other matrix representations that require less hardware. In the end, this question is also solved by trial and error with different percentages of unstructured and structured pruning.

\section{Experimental Set-up, Challenges and Results}
\label{sec:challenges_results}
For the experiments in the paper, we consider a small quadruped robot called Bittle. Bittle is a low-resource robotic platform that lacks positional encoders for its servos, and hence most reward functions used for imitation learning do not work on it.  Also, it has limited memory and computational capabilities, making it the perfect subject to test our method. In the following subsection we show how decision transformers can be applied to quadruped locomotion for Bittle in simulation to produce natural looking gaits.

\subsection{Expert demonstrations and custom reward structure}
\begin{figure}
    \centering
    \includegraphics[width=0.45\textwidth]{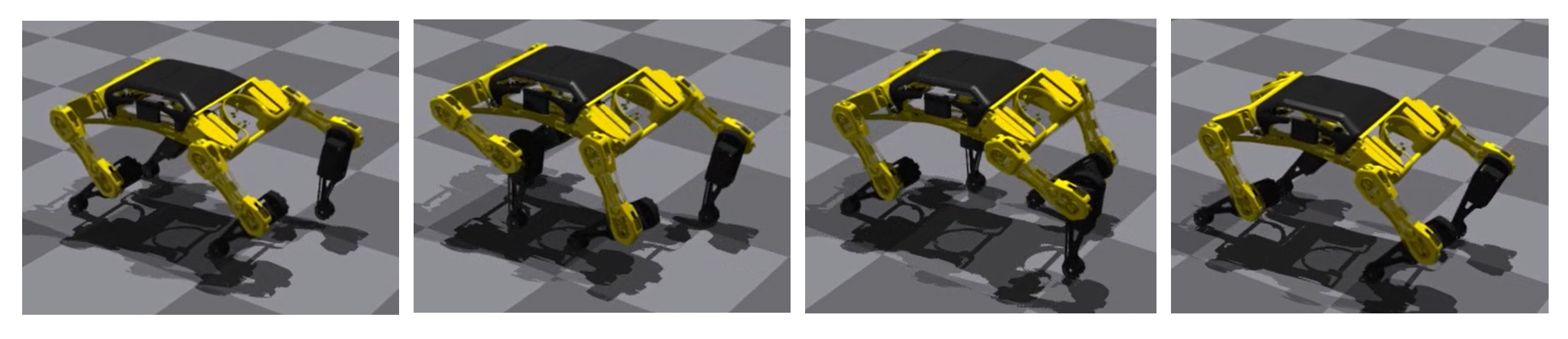}
    \caption{Example gait for Bittle in simulated Isaac Gym environment}
    \label{fig:example_gait}
\end{figure}

We simulated Bittle using Isaac Gym and the URDF suggested in Bittle's official documentation. We specified a stiffness of $5.25$ and a damping factor of $0.25$ for all the servos in the robot. We also changed the effort characteristic of all the joints to $5$. We generated the expert demonstrations by running a kinematics module that calculated potentially useful gaits using inverse kinematics and a desired linear velocity of Bittle's body. We  manually selected two gait options from the pool of six possible gaits based on their stability and their visual similarity to Bittle's pre-defined gaits. We simulated $1500$ different environments with different initial positions for the robot, different noise factors, and random disturbances on the dynamic and static friction. We collected sequences of states-actions for each trajectory. The states for this setup are composed of the previous action and the readings from the only available sensor, Bittle's IMU, which will give us access to the linear and angular velocity, and acceleration of Bittle's body. We used these sequences of states and actions to generate rewards $r_t$ for each time $t$ using the following formula:
\begin{flalign}
    r_t = 2*v_{xy}(t) - 1*v_{z}(t) - 0.05*\omega_{xy}(t) \nonumber \\ + 
    0.5 \omega_{z}(t) -0.01*\delta_a(t)
\end{flalign}
Where $v_{xy}(t)$ corresponds to the linear velocity in the $xy$ direction as measured by Bittle's IMU at time $t$. $v_{z}(t)$ corresponds to the linear velocity in the $z$ direction as measured by Bittle's IMU at time $t$. $\omega_{xy}(t)$ corresponds to the angular velocity in the $xy$ direction as measured by Bittle's IMU. $\omega_{z}(t)$ corresponds to the angular velocity in the $z$ direction as measured by Bittle's IMU, and $\delta_a$ corresponds to the $l_2$ norm of the difference between the previous action $a_{t-1}$ and the current action $a_t$. Once we have augmented all the trajectories with their respective reward signals, we use the trajectories to train the decision transformer. We evaluate the trained decision transformer on $250$ randomly generated environments and obtain an average reward of $196.5$. An example of frames for the visual rendition of the trajectory can be found in Fig. \ref{fig:example_gait}.

\subsection{Model Compression Analysis}

\begin{figure}
    \centering
    \includegraphics[width=0.45\textwidth]{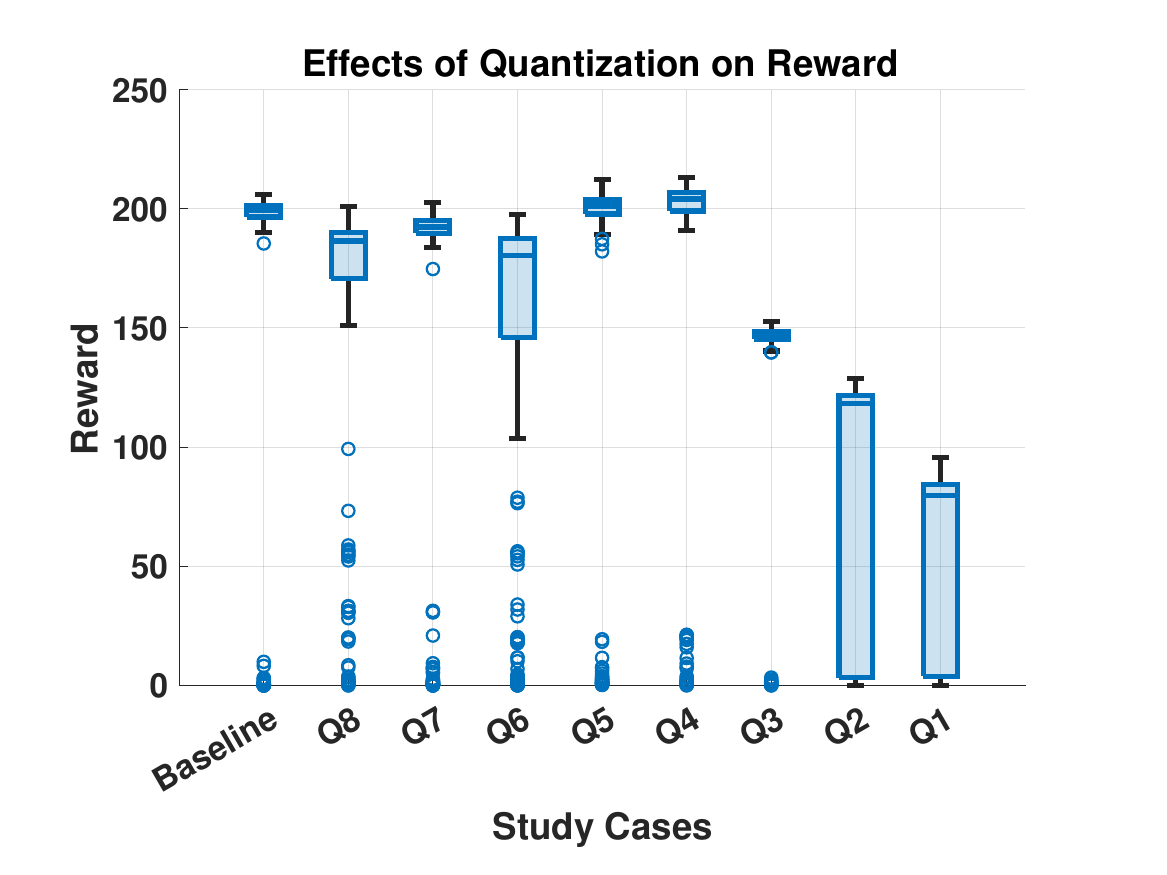}
    \caption{Statistics of the reward obtained after quantization for different numbers of bits, based on 250 Monte-Carlo points.}
    \label{fig:quant_reward}
\end{figure}

Considering the resource-constrained environment of robotic platforms discussed in this work, we applied quantization and pruning techniques to minimize the size of our decision transformer. Given the inherent randomness of conventional simulation environments for robotic platforms, we validate our compression approaches based on Monte-Carlo simulations for 250 points. In this section, we analyze the effect that every strategy has on the simulated model reward to contrast it against the size reduction achieved. Along with checking the statistics of the obtained reward for each study case, we also verified the robotic platform video simulations to confirm natural gaits and correct movement. For the following figures, Q stands for quantization along with the targeted number of bits, P stands for pruning, P+Q is a combination of first pruning and then quantizing (Q+P is the opposite order), and FT stands for the fine-tuned model.

As Fig. \ref{fig:quant_reward} summarizes, we implemented uniform quantization for different targeted numbers of bits, ranging from 8 down to 1. Our results and robotic platform video simulations suggest an acceptable reward loss down to 4-bits quantization. Interestingly, 4-bit quantization shows similar data to the baseline full-precision case. In both study cases, there are two well-defined data clusters around 200 and below 20, and the median is statistically equivalent despite the higher number of outlier cases after quantization. The portion with a significant number of small rewards can be justified by bad initialization from the robotics simulation environment. The quantization results presented can still be extended to obtain more accurate data distribution metrics, however, considering the long simulation times we constrained the number of experiments to 250 for all our analyses.

Quantization techniques typically enable massive model size reduction compared to other techniques like pruning. Therefore, considering our goal of minimizing the decision transformer size, we decided to apply pruning on top of our uniform quantization algorithm. In this case, we study the reward loss for only quantization, quantization plus pruning in different orders, and the effect of fine-tuning after pruning with subsequent quantization. For the following results, we applied a pruning method balancing unstructured and structured procedures as described in Section \ref{sec:problem_formulation}.

\begin{figure}
    \centering
    \includegraphics[width=0.49\textwidth]{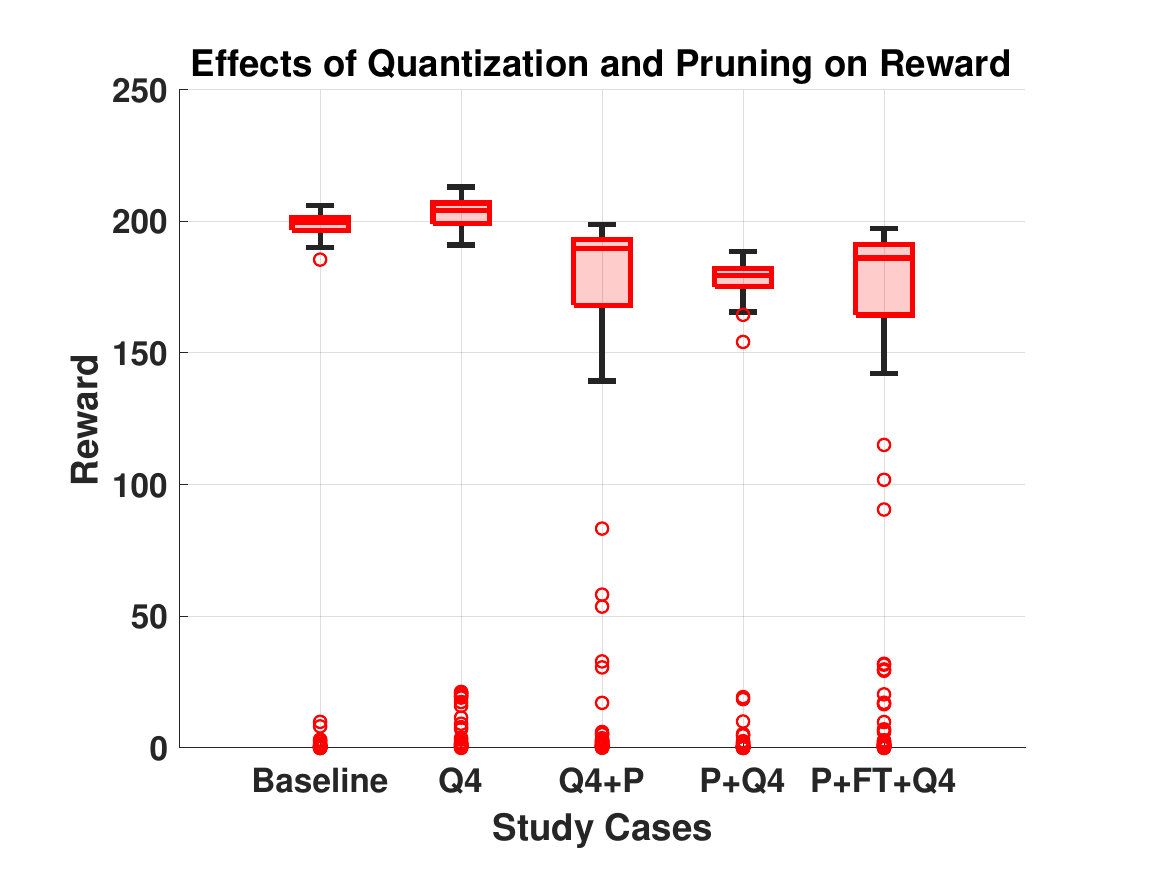}
    \caption{Statistics of the reward obtained for different combinations of quantization and pruning strategies based on 250 Monte-Carlo points.}
    \label{fig:quant_prun_reward}
\end{figure}

In Fig. \ref{fig:quant_prun_reward} we studied the combination of quantization and pruning, and its effect on our reward compared to the full-precision baseline and only-quantization cases. Here we also analyze how post-pruning fine-tuning may contribute to an improvement in the reward. Our results suggest that pruning has a more notorious impact on the reward median than only quantization. Also, in our robotic platform application, quantization before pruning performs better in terms of reward median than in the opposite case. However, we also illustrate how an additional post-pruning fine tune of 20\% of the initial training workload can improve the results for pruning before quantization.

\begin{figure}
    \centering
    \includegraphics[width=0.49\textwidth]{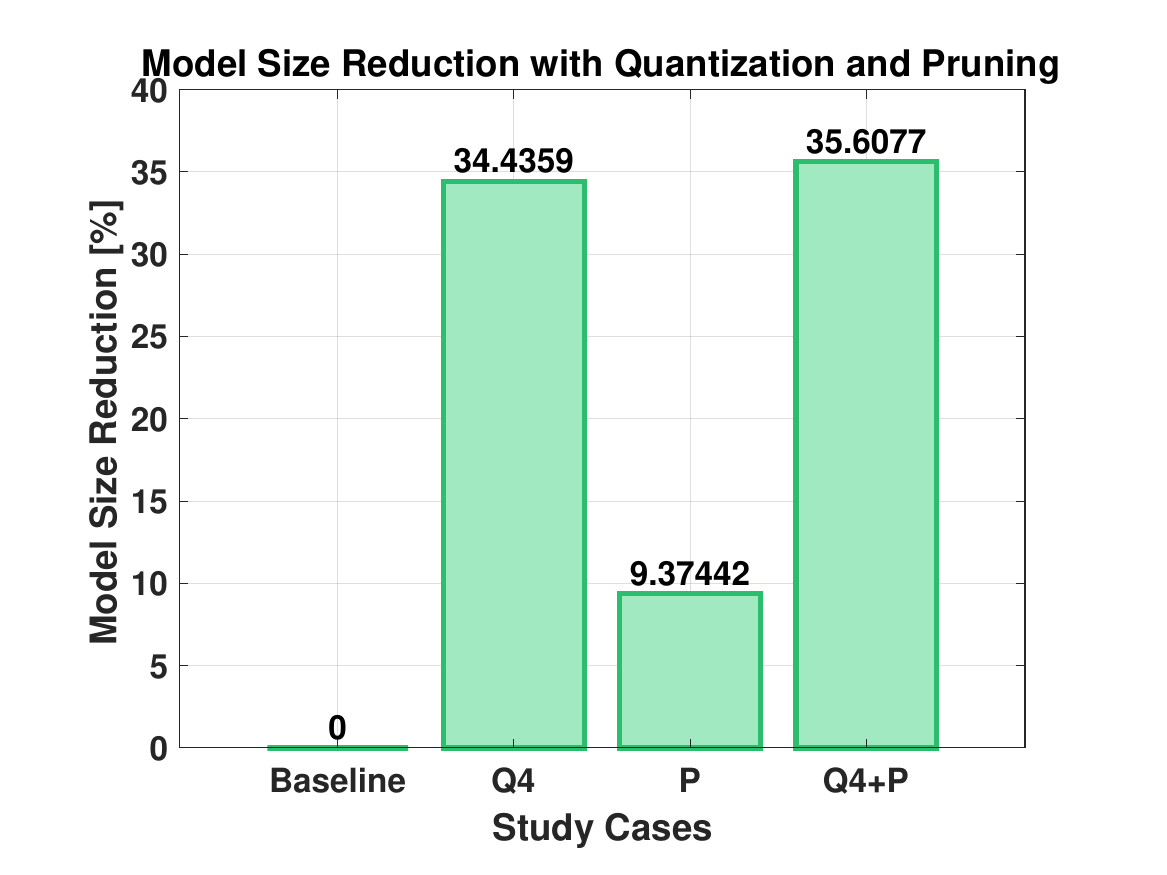}
    \caption{Normalized model size reduction across the different compression techniques applied.}
    \label{fig:size_reduction}
\end{figure}

Given the need to deploy our decision transformer model into the actual robotic platform hardware, we must examine the model size reduction enabled by the compression techniques applied in this work. Fig. \ref{fig:size_reduction} depicts the percentage of model size reduction regarding the full-precision baseline for each compression technique. Here, \textit{Q4+P} covers all the combinations between quantization and pruning.

We base our model size reduction analysis on the number of parameters of our decision transformer. As previously explained in Section \ref{sec:problem_formulation}, we applied quantization and pruning techniques exclusively on the fully connected and convolutional layers of our model. Therefore, any size reduction we can account for will be related to lower precision of the parameters (quantization), or parameters removed from the model (pruning). Finally, we calculate a size reduction percentage with respect to the size of the actual model file that would be stored in the hardware platform. Thus, for better generality of our estimations, we take into account information that may be stored in the model file for operating system compatibility or may not necessarily be related to the decision transformer. For instance, different file formatting can output lower file sizes (in bytes). The standard output used in PyTorch is Pickle, and we based our calculations on this extension.

As Fig. \ref{fig:size_reduction} shows, the most aggressive size reduction can be obtained by employing only quantization techniques. On the other hand, we could only achieve a maximum of around 9\% file size reduction through pruning without any considerable impact on the reward. Therefore, when applying pruning on top of 4-bit quantization, the model size reduction related to pruning becomes almost negligible, considering that we are removing 4-bit parameters instead of 32-bit ones.

\section{Discussion and Future Work}
\label{sec:dicussion_future_work}

In terms of future work, we aim to specify a better reward structure -- independent of the kinematics of the system to reduce computational time. We also want to deploy the tiny decision transformer to a real Bittle using a Raspberry PI to validate our results in hardware experiments. We will measure the latency of the sequence predictions and check its effect on the resulting gait. 

From the model compression perspective, more complex quantization techniques such as term quantization \cite{kung2020} and piece-wise linear quantization \cite{pwlq} can be explored. These techniques showed stronger model reductions with negligible accuracy drops in models based exclusively on fully connected and convolutional layers. We would like to address these two techniques on transformer models to possibly achieve better performance. There are also transformer-specific pruning techniques that we can incorporate to have a better decision metric when choosing lower-significant connections. For instance, we can utilize previously studied methods to efficiently prune heads \cite{NEURIPS2019_2c601ad9, voita-etal-2019-analyzing}. Furthermore, we can add knowledge distillation, Huffman encoding, and other techniques to amplify our model compression pipeline.





\bibliography{main}
\bibliographystyle{mlsys2021}



\end{document}